\documentclass[letterpaper, 10 pt, conference]{ieeeconf}  
\usepackage{amsmath}
\usepackage{tensor}
\usepackage{bm}
\usepackage{amssymb}
\usepackage{graphicx}
\usepackage{caption}
\usepackage{color}
\usepackage{midfloat}
\usepackage{float}
\usepackage{rotating}
\usepackage{multirow}
\usepackage{hhline}
\usepackage{booktabs}
\usepackage{makecell}
\usepackage{longtable}
\usepackage{array}

\usepackage{paralist}
\usepackage{multicol}
\usepackage{chngcntr}
\usepackage{arydshln,leftidx,mathtools}
\usepackage{threeparttable}
\usepackage{tabularray}

\IEEEoverridecommandlockouts                              

\title{\LARGE \bf
Origami-Inspired Soft Gripper with Tunable Constant Force Output
}

\author
{Zhenwei Ni, Chang Xu, Zhihang Qin, Ceng Zhang, Zhiqiang Tang, Peiyi Wang and Cecilia Laschi
\thanks{This research is partly supported by NUS Start-up Grant "RoboLife". This research is also supported by the National Research Foundation (NRF), Prime Minister’s Office, Singapore under its Campus for Research Excellence and Technological Enterprise (CREATE) programme. The Mens, Manus, and Machina (M3S) is an interdisciplinary research group (IRG) of the Singapore MIT Alliance for Research and Technology (SMART) centre. (\textit{Corresponding author: Peiyi Wang})}
\thanks{Zhenwei Ni, Chang Xu, Zhihang Qin, Ceng Zhang, Peiyi Wang and Cecilia Laschi are with Department of Mechanical Engineering and Advanced Robotics Centre, National University of Singapore, Singapore, Singapore}
\thanks{Zhenwei Ni and Zhiqiang Tang are with Singapore MIT Alliance for Research and Technology (SMART) centre, Singapore, Singapore}
}

\begin{document}
\maketitle
\thispagestyle{empty}
\pagestyle{empty}

\begin{abstract}

Soft robotic grippers gently and safely manipulate delicate objects due to their inherent adaptability and softness.
Limited by insufficient stiffness and imprecise force control, conventional soft grippers are not suitable for applications that require stable grasping force.
In this work, we propose a soft gripper that utilizes an origami-inspired structure to achieve tunable constant force output over a wide strain range. The geometry of each taper panel is established to provide necessary parameters such as protrusion distance, taper angle, and crease thickness required for 3D modeling and FEA analysis. Simulations and experiments show that by optimizing these parameters, our design can achieve a tunable constant force output. 
Moreover, the origami-inspired soft gripper dynamically adapts to different shapes while preventing excessive forces, with potential applications in logistics, manufacturing, and other industrial settings that require stable and adaptive operations.

\end{abstract}


\section{INTRODUCTION}
Soft mechanical structures attract significant attention for their high compliance, adaptability, and energy dissipation capabilities. In biomedical engineering, they enable wearable sensors~\cite{lin2022soft_wearableSensor} and implantable devices~\cite{lacour2016materials_implantableDevice} that conform to the human body. In industrial applications, these structures are employed in energy-absorbing systems such as impact protection layers~\cite{kumar2023high_softLayer} and damping structures~\cite{shu2022viscoelastic_softDamper}.

\begin{figure}[!t]
    \centering
    \includegraphics[width=1\columnwidth]{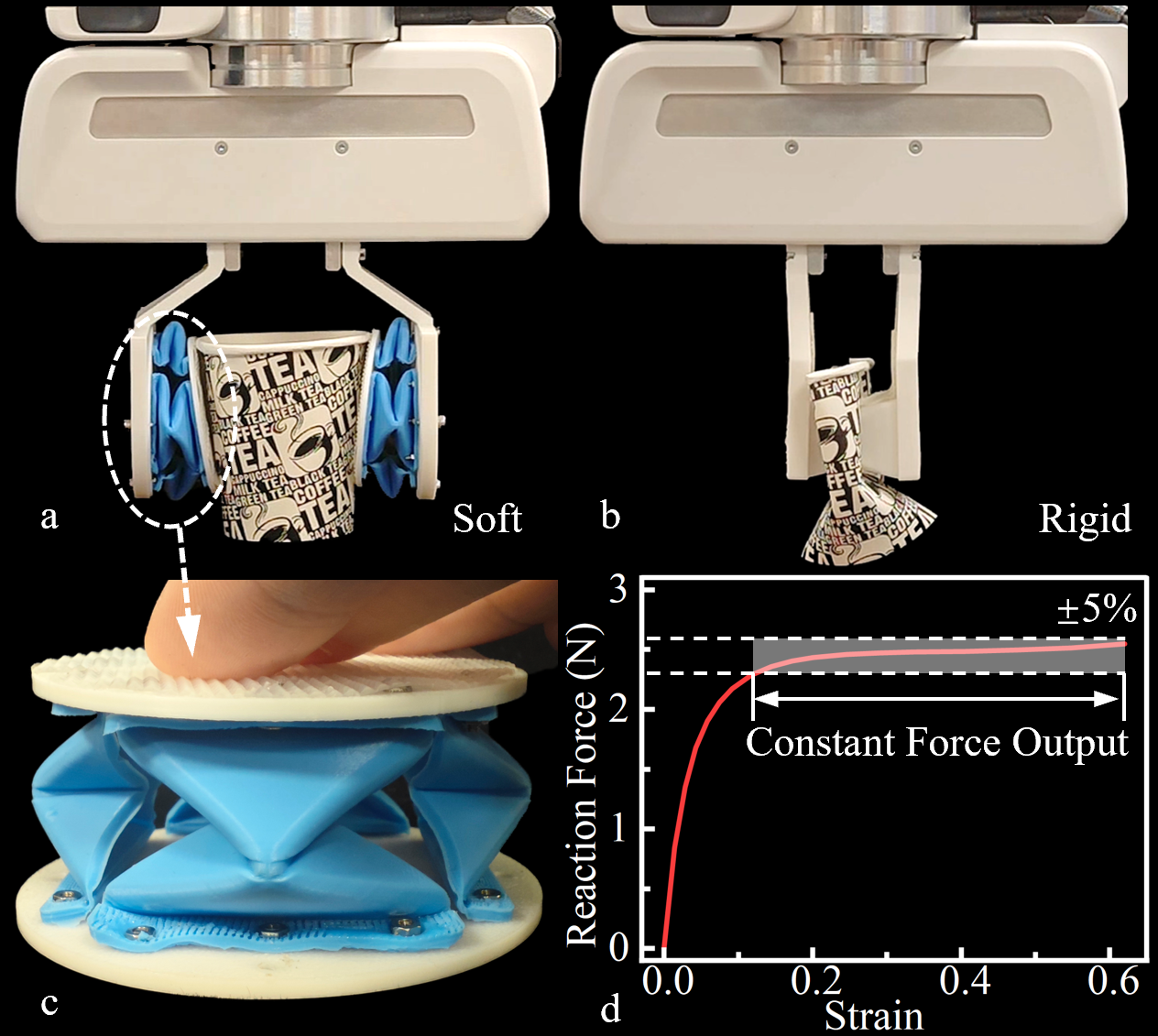} 
    \caption{ Grasping a paper cup using (a) an origami-inspired soft gripper and (b) a rigid gripper.
    (c)~Deformable modular gripper and (d) the corresponding reaction force–strain property (gentle and constant output force).
    }
    \label{fig:graphical_abstract}
\end{figure}

Beyond these domains, soft mechanical structures have also found applications in robotic grippers. Unlike traditional rigid grippers, soft grippers leverage the inherent compliance of their materials to grasp irregularly shaped, fragile, or delicate objects without causing damage. Pneumatic soft robotic grippers, one of the most classic soft gripping methods, use inflatable elastomeric chambers for adaptive and gentle grasping~\cite{hao2016universal,ye2022design}. Grippers with electroactive polymer actuators (EAP), including dielectric elastomer actuators (DEA)~\cite{gupta2019soft_DEAs,shian2015dielectric_DEAs} and ionic polymer-metal composites (IPMC)~\cite{he2020highly_IPMCs}, offer another flexible approach, utilizing electrically responsive materials to achieve efficient and precise gripping. Some researchers have also focused on combining Shape Memory Materials with soft grippers~\cite{schonfeld2021actuating_SMMsHeat,liu2020novel_SMMsHeat}. 

While most soft grippers excel in tasks requiring compliance and adaptability, the lack of stable manipulation with constant force grasping under dynamic conditions limits their applications in real scenarios such as factory automation. Additionally, common soft gripper materials such as silicone and EAP exhibit nonlinear mechanical behavior, including phenomena like hysteresis~\cite{soft_mat_hysteresis} and viscoelasticity~\cite{soft_mat_viscoelasticity_2}, which in turn complicates precise force control and repeatability. These limitations highlight the need for innovative designs that combine the compliance and adaptability of soft grippers with stable mechanical output. 

Origami-inspired structures have emerged as a key paradigm in engineering due to their unique geometric and mechanical properties, such as high strength to weight ratio~\cite{ori_high_ratio}, modularity~\cite{ori_modular}, multi-modality \cite{wang2023tristable}, predictable motion mechanisms~\cite{ori_motion_predictable}, and energy absorption capabilities~\cite{ori_energy_absorption}. Many studies have explored the integration of origami-inspired structures with robotic grippers~\cite{soft_robot_grippers}. By leveraging nonlinear deformation and bistable configurations, some studies achieve rapid actuation~\cite{rapid_response_ori_gripper1,rapid_response_ori_gripper2}. Other focuses on alternative structural principles to enable precise and modular grasping mechanisms~\cite{kirigami_grasping}. Moreover, the incorporation of advanced sensing~\cite{sensing_ori_gripper} and intelligent control strategies~\cite{control_strategies_ori_gripper} further enhances performance by providing real-time feedback and improved precision. Although these studies have demonstrated impressive advances in rapid actuation, morphological adaptability, and integrated control, the potential of origami-inspired grippers to deliver a constant force output remains to be conclusively proven. 

In this work, we propose an origami-inspired soft structure with a wide range of constant force output and demonstrate its application in grippers (Fig. \ref{fig:graphical_abstract}). The primary structure consists of a module composed of four basic origami unit panels (hereafter referred to as \textit{panel}) arranged in a circular pattern, with the panels inspired by the classic Waterbomb pattern. The Waterbomb pattern features a distinctive folding design that allows for significant deformation while preserving structural integrity. Unlike the traditional Waterbomb structure, our tapered panel incorporates sixfold Waterbomb inlays. Studies have shown that this structure exhibits unique folding behavior by mitigating interlayer motion coupling, enabling independent layer deformation, and enhancing structural adaptability and mechanical diversity~\cite{unique_waterbomb}. The force response curve of this structure was analyzed from an elastic mechanics perspective in \cite{unique_waterbomb_force_analysis}, providing the theoretical foundation and inspiration for achieving constant force output in our gripper design. Compared to conventional origami-inspired grippers, our proposed design reduces the soft gripper's demand for precise control while ensuring a consistent and stable force output during compression. We demonstrate that the structure and gripper achieve a desired constant force output by using different soft materials and adjusting design parameters.

\section{Design and Simulation }

\subsection{Origami-inspired Geometry.} 

The panel serves as the fundamental folding element for each module. The geometry of each panel is shown in Fig. \ref{fig:geometric design}, featuring an origami-inspired configuration.  
Six boundary points,
A, A$_1$, B, B$_1$, C, C$_1$, and O, are defined in this trapezoidal origami structure.
M and N are the midpoints of lines AA$_1$ and BB$_1$, respectively.

The positions of point C, C$_1$ and O are represented by $\gamma_1, \gamma_2, \beta_1, \beta_2$. When the panel is fully folded, the relationships between these angles are:
\begin{equation}
    \gamma_1 - \gamma_2 = \beta_1 - \beta_2,
\end{equation}
\begin{equation}
    \gamma_1 + \gamma_2 + \beta_1 + \beta_2 = \pi,
\end{equation}
Then $\gamma_1$ and $\gamma_2$ can be derived:
\begin{equation}
    \gamma_1 = \frac{\pi}{2} - \beta_2, \quad \gamma_2 = \frac{\pi}{2} - \beta_1.
\end{equation}
To avoid interference between crease $\mathrm{OC}$ and $\mathrm{OC}_1$, the condition \( \gamma_2 < \beta_2 \) must be satisfied. Consequently, 
\begin{equation}
    \beta_1 + \beta_2 > \frac{\pi}{2}
\end{equation}
Here, we introduce necessary parameters to describe the origami-inspired structure.

\subsubsection{Taper Angle \( \alpha \)}

\begin{figure}[!t]
    \centering
    \includegraphics[width=1\columnwidth]{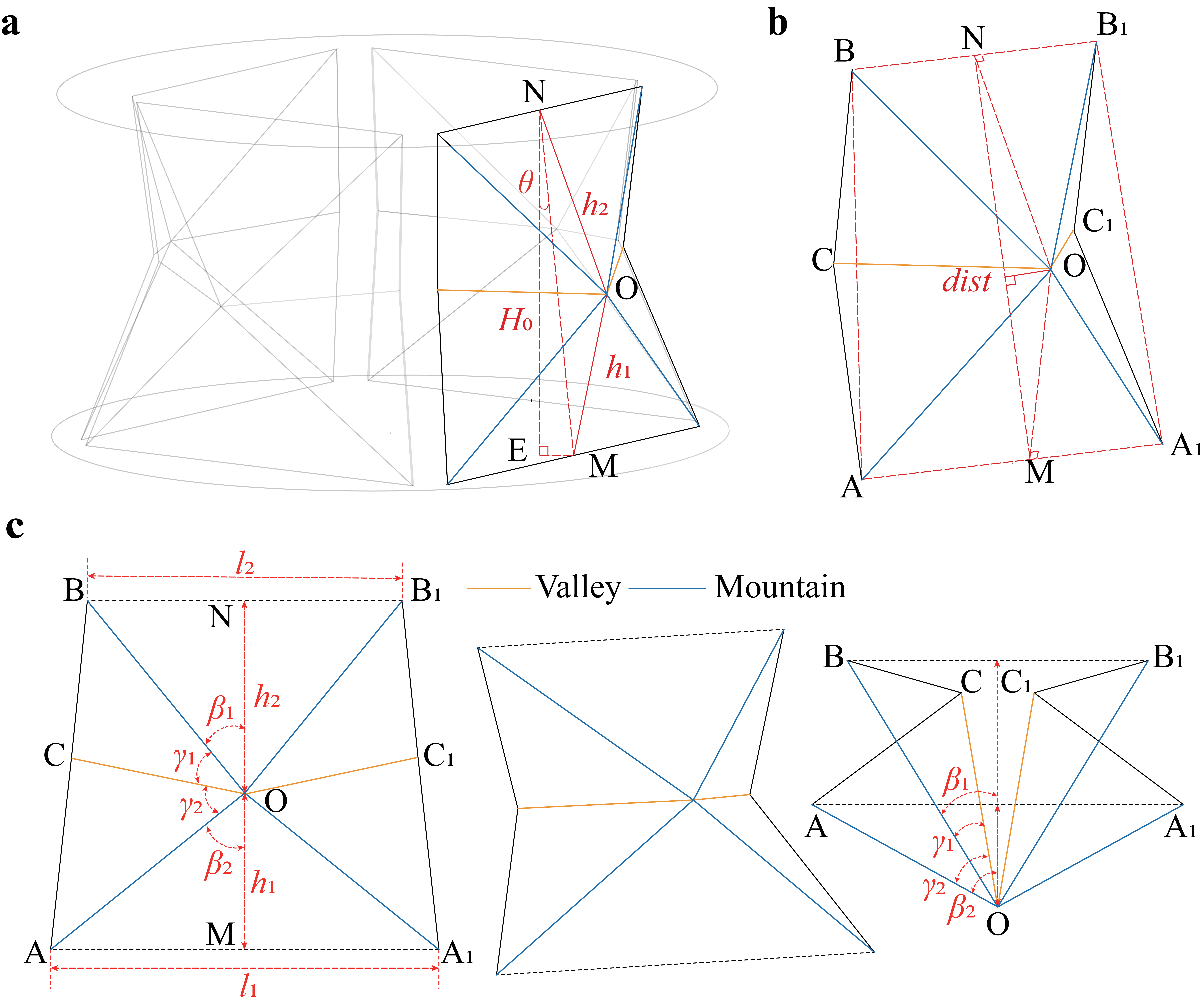}
    \caption{Scheme of the origami-inspired structure and folding mode.~(a)~Geometry of single origami panel in each module.~(b)~Definition of the protrusion distance $dist$.
    ~(c)~Planar geometry and folding pattern}
    \label{fig:geometric design}
\end{figure}

The taper angle \( \alpha \) is a key parameter that characterizes the overall tapering of a conical module. In Fig. \ref{fig:geometric design}a, the angle \( \theta \) represents the inclination of each panel. Consequently, the taper angle  \(\alpha = 2\theta\). Notably, \(\alpha=0\) yields a cylindrical configuration.

\subsubsection{Protrusion Distance \(\mathit{dist}\)}
The protrusion distance \(\mathit{dist}\) is defined as the Euclidean distance from point \(\mathrm{O}\) to the plane \(\mathrm{AA}_1\mathrm{BB}_1\). From geometric symmetry, we have \(\mathrm{OM} \perp \mathrm{AA}_1\) and \(\mathrm{ON} \perp \mathrm{BB}_1\). Since \(\mathrm{AA}_1\) and \(\mathrm{BB}_1\) are parallel, it follows that both \(\mathrm{AA}_1\) and \(\mathrm{BB}_1\) are parallel to the plane \(\mathrm{OMN}\). Consequently, the plane \(\mathrm{AA}_1\mathrm{B}_1\mathrm{B}\) is parallel to \(\mathrm{OMN}\). Therefore, the perpendicular drawn from point \(\mathrm{O}\) to the line \(\mathrm{MN}\) has a length equal to \(\mathit{dist}\). Given that \(\mathrm{OM} = h_1\), \(\mathrm{ON} = h_2\), and \(\mathrm{MN} = \frac{H_0}{\cos\theta}\) (where \(H_0\) is the initial height of the structure), \(\mathit{dist}\) can be computed using Heron’s formula:
\begin{equation}
    \mathit{dist} 
    = 2 \cdot \frac{\sqrt{p \bigl(p - h_1 \bigr) \bigl(p - h_2 \bigr) \bigl(p - H_0/\cos\theta\bigr)}}{H_0/\cos\theta},
    \label{eq:dist}
\end{equation}
where
\begin{equation}
    p = \frac{h_1 + h_2 + H_0/\cos\theta}{2}.
\end{equation}

\subsubsection{Height Ratio}

Point \(\mathrm{O}\) moves along the segment connecting points \(\mathrm{M}\) and \(\mathrm{N}\). We define the height ratio as
\begin{equation}
    t = \frac{l_{\mathrm{ON}}}{l_{\mathrm{OM}}} = \frac{h_2}{h_1}.
\end{equation}

Each module achieving fully folded state must stratify the following geometric relationship:
\begin{equation}
    l_{\mathrm{EM}} = h_2 - h_1 = H_0 \tan\theta,
    \label{eq:l_EM}
\end{equation}
where \(H_0\) is a given parameter. Based on Equations \eqref{eq:dist} and \eqref{eq:l_EM}, the parameters \(h_1\) and \(h_2\) can be expressed as
\begin{equation}
    h_1 = \frac{2\textit{dist}^2 + H_0^2(1 - \sin\theta)}{2H_0 \cos\theta},
\end{equation}
\begin{equation}
    h_2 = \frac{2\textit{dist}^2 + H_0^2(1 + \sin\theta)}{2H_0 \cos\theta}.
\end{equation}
Therefore, the height ratio \( t \) is given by

\begin{equation}
    t = 1 + \frac{2H_0^2\sin\theta}{2\,\textit{dist}^2 + H_0^2\left(1 - \sin\theta\right)}.
\end{equation}

\subsection{Panel Modeling}

\begin{figure*}[!t]
    \centering
    \includegraphics[width=0.9\textwidth]{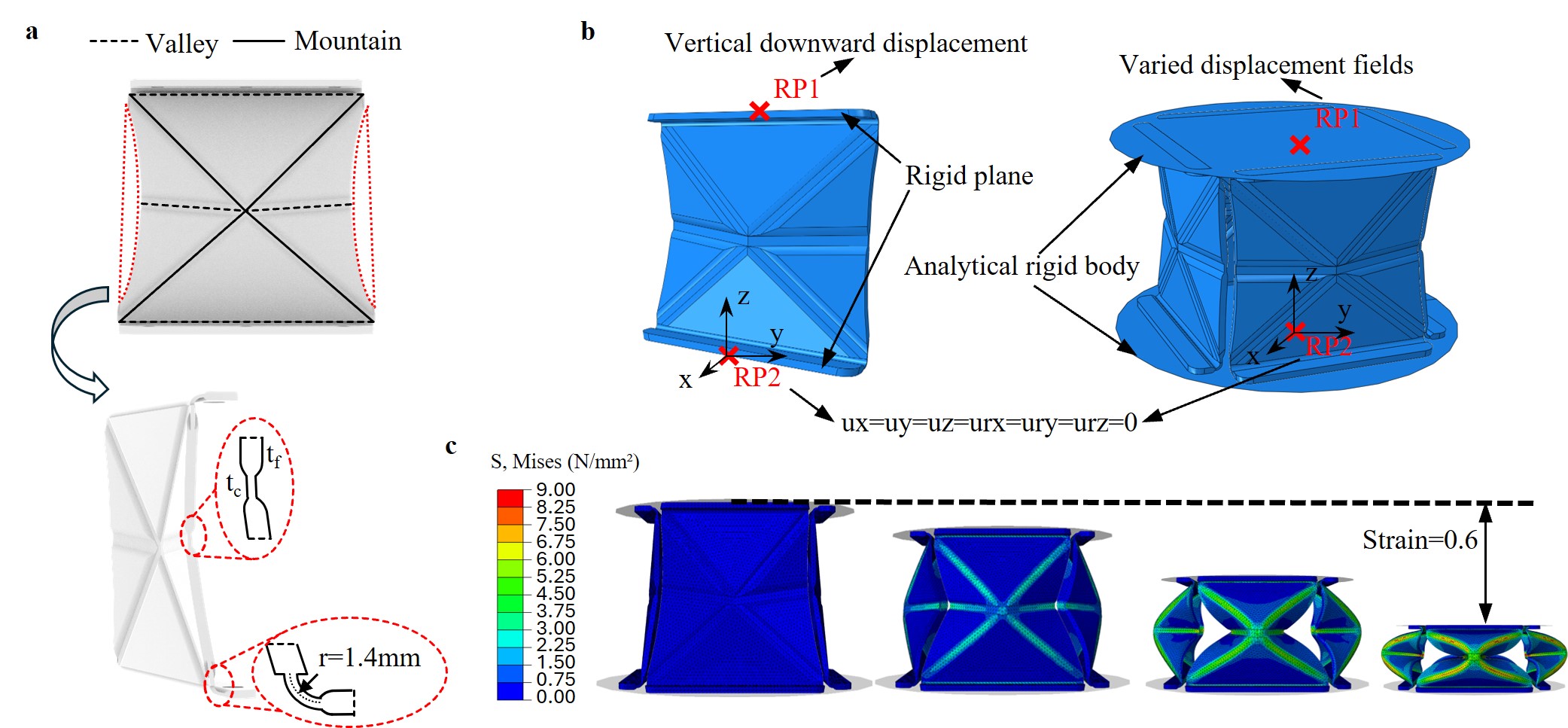}
    \caption{Modeling and simulations.~(a)~3D modeling of the panel.~(b)~Simulation setup for the panel and module.~
    (c)~Numerically obtained deformation processes of the module in post-buckling analysis.}
    \label{fig:fea}
\end{figure*}

The 3D modeling of the origami-inspired panel was completed in \textit{SolidWorks} as shown in Fig. \ref{fig:fea}a. The key common design parameters are:  
the lower base length (\( l_1 \)) is 50 mm,  
while the upper base length (\( l_2 \)) is 45 mm.  
The module height (\( H_0 \)) is 43 mm,  
and the facet thickness (\( t_f \)) is 1.5 mm.  The crease regions were designed with a thickness smaller than that of the facets. The top and bottom of the panel are with a thickness of 1.5 mm and a bending radius of 1.4 mm to ensure stability and controlled deformation while providing boundary support and allowing free movement. A series of samples with varying geometric configurations (Table. \ref{tab:PanelModelParameters}) were designed and used in subsequent parameter and simulation analysis.

\begin{table}
\centering
\caption{Panel Model Parameters}
\label{tab:PanelModelParameters}
\resizebox{1\linewidth}{!}{
\begin{tblr}{
  column{2} = {c},
  column{3} = {c},
  column{4} = {c},
  column{5} = {c},
  column{7} = {c},
  column{8} = {c},
  column{9} = {c},
  column{10} = {c},
  cell{13}{1} = {c=5}{},
  hline{1,14} = {-}{0.08em},
  hline{2} = {-}{},
}
\textbf{Type} & {\textbf{$dist$ }\\\textbf{ (mm)}} & {\textbf{$\alpha$ }\\\textbf{ (°)}} & {\textbf{$t_c$ }\\\textbf{ (mm)}} & \textbf{$n^*$} & \textbf{Type}                                      & {\textbf{$dist$ }\\\textbf{ (mm)}} & {\textbf{$\alpha$ }\\\textbf{ (°)}} & {\textbf{$t_c$ }\\\textbf{ (mm)}} & \textbf{$n^*$} \\
M1           & 0.0                              & 0                                   & 0.54                              & 1                                      & M12                                                 & 4.0                              & 6                                   & 0.54                              & 1                                      \\
M2            & 1.0                              & 0                                   & 0.54                              & 1                                      & M13                                                 & 4.0                              & 6                                   & 0.68                              & 1                                      \\
M3            & 1.5                              & 0                                   & 0.54                              & 1                                      & M14                                                 & 4.0                              & 6                                   & 0.81                              & 1                                      \\
M4            & 2.0                              & 0                                   & 0.54                              & 1                                      & M15                                                 & 4.0                              & 6                                   & 0.95                              & 1                                      \\
M5            & 3.0                              & 0                                   & 0.54                              & 1                                      & M16                                                 & 4.0                              & 6                                   & 1.08                              & 1                                      \\
M6            & 4.0                              & 0                                   & 0.54                              & 1                                      & M17                                                 & 2.0                              & 10                                  & 0.54                              & 1                                      \\
M7            & 4.0                              & 0                                   & 0.54                              & 1                                      & M18                                                 & 2.0                              & 10                                  & 0.54                              & 0.84                                   \\
M8            & 4.0                              & 4                                   & 0.54                              & 1                                      & M19                                                 & 2.0                              & 10                                  & 0.54                              & 0.71                                   \\
M9            & 4.0                              & 8                                   & 0.54                              & 1                                      & M20                                                 & 2.0                              & 10                                  & 0.54                              & 0.60                                   \\
M10           & 4.0                              & 12                                  & 0.54                              & 1                                      & M21                                                 & 2.0                              & 10                                  & 0.54                              & 0.51                                   \\
M11           & 4.0                              & 16                                  & 0.54                              & 1                                      & M22                                                 & 2.0                              & 10                                  & 0.54                              & 0.43                                   \\
\textbf{$n^*$}: scaling ratio &                                  &                                     &                                   &                                     
\end{tblr}
}
\end{table}

\subsection{Finite Element Analysis Setup}
The origami-inspired structures, characterized by their complex geometry and high flexibility, often exhibit significant nonlinearity and large deformations that conventional static methods fail to capture accurately. Therefore, a buckling and post-buckling analysis in ABAQUS was employed in this study to investigate the deformation behavior. Critical loads and instability modes are determined in the buckling analysis. The scaled results are then used as initial imperfections in post-buckling analysis to simulate deformation beyond buckling and reveal performance characteristics.

Each panel in Table \ref{tab:PanelModelParameters} was assigned a 3D solid section with material properties defined as TPU95A (Young’s modulus: 26 MPa, Poisson’s ratio: 0.4). The upper and lower connecting structures were modeled as rigid planes, with reference points located at their respective centers. To simulate the compression process, the lower reference point was constrained with an encastre boundary condition (fixing all six degrees of freedom), while a vertical downward displacement was applied to the upper reference point. The mesh was generated using C3D4 elements, with a global element size 1.0. For more accurate stress analysis in the crease regions, the element size was refined to 0.8 in those areas. 
Buckling analysis was conducted using the \textit{Buckle} step. The first three buckling modes from the buckling analysis are shown in Fig. \ref{fig:fea}b. The first buckling mode, which is the most probable to occur under compression, was selected as the geometric imperfection for the post-buckling analysis. The post-buckling analysis was then carried out using the \textit{Riks}.

We also conducted FEA on the module for further experimental data validation (Fig. \ref{fig:fea}c), which shares similar FEA settings with the panels. Specifically, four panels were arranged in a circular array and sandwiched between two disks, with each panel’s bottom center located 30 mm from the module’s central axis. The disks were modeled as analytical rigid bodies with central reference points, and each panel was defined as an independent assembly sharing a common part definition. Panels and disk surfaces were bonded using tie constraints. The module was analyzed using the same buckling and post-buckling approach as the single panel, with the bottom disk's reference point constrained by boundary conditions and the upper disk subjected to varied displacement fields to simulate different compression modes.

\subsection{Simulation Analysis of Panel Parameters}
\subsubsection{Protrusion distance dist}

\begin{figure*}[!t]
    \centering
    \includegraphics[width=0.81\textwidth]{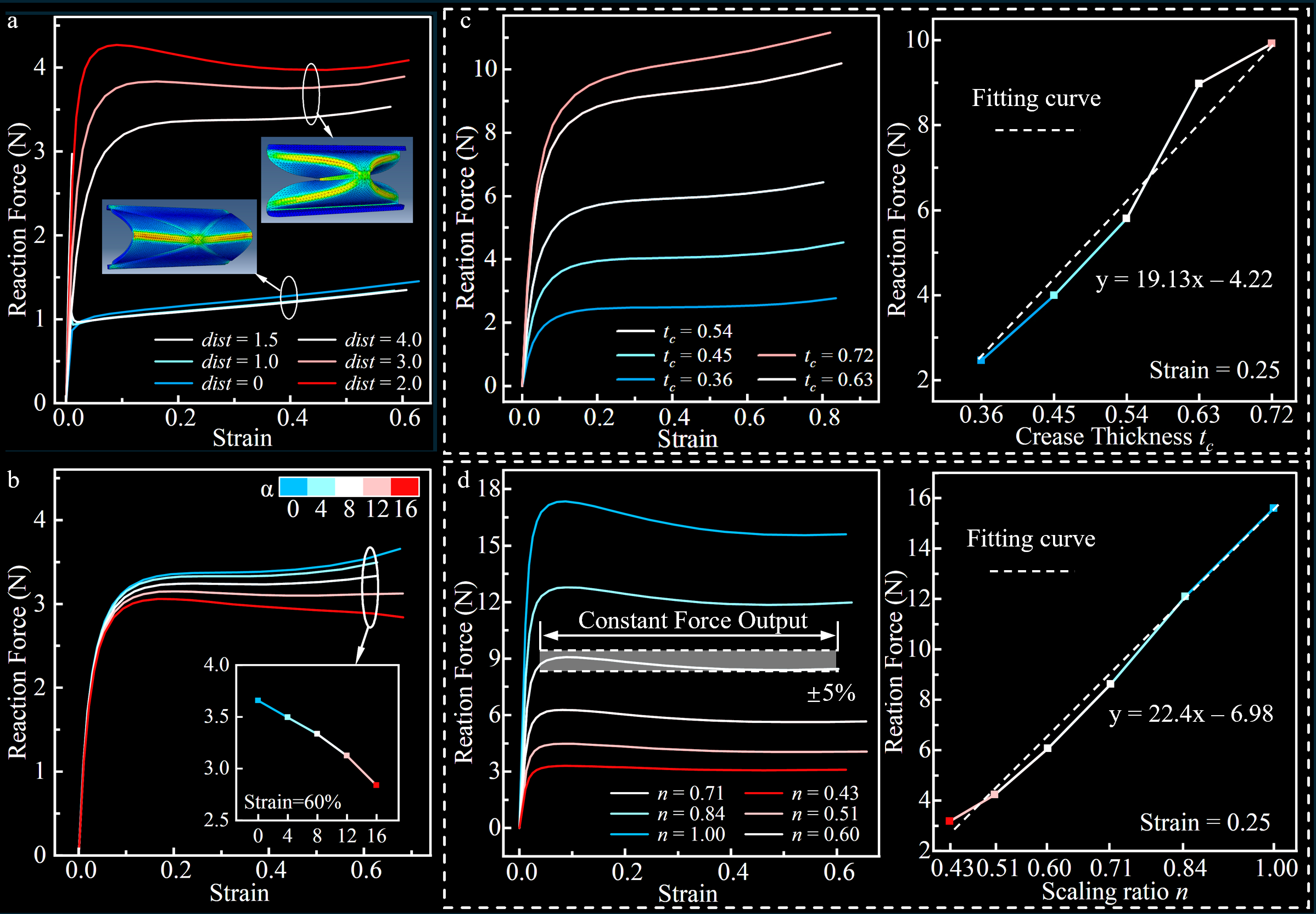}
    \caption{Reaction Force-Strain analysis for different structure parameters, including (a) protrusion distance \textit{dist}, (b) taper angle \textit{$\alpha$}, (c) crease thickness \textit{$t_c$ }, and (d) scaling ratio \textit{$n$}.}
    \label{fig:simulation results}
\end{figure*}

The Reaction Force (RF)-strain curves for panels with different protrusion distances M1 to M6 are obtained (Fig. \ref{fig:simulation results}a). Results reveal that the protrusion distance \textit{dist} of the panel causes different deformable states, including target folding, unidirectional, and one-sided folding mode under compression.
When \textit{dist} is very small (approaching zero), the panel behaves almost like a 2D flat structure, making it prone to bending asymmetrically to one side rather than folding uniformly. However, when \textit{dist} exceeds 2 mm, the panel transitions into the targeted folding mode. We also found that the target folding provides a nearly constant reaction force, which is significantly higher than that observed in the one-sided folding mode. Meanwhile, increasing \textit{dist} can reduce the magnitude of the output force.

\subsubsection{Taper angle \text{$\alpha$}}
We simulated panels with \textit{dist} of 4 mm under different \text{$\alpha$} (M7 to M11), and their RF-Strain curves are shown in Fig. \ref{fig:simulation results}b. When the taper angle $\text{$\alpha$} = 0$, the RF-Strain curve shows the highest reaction force, indicating that the panel provides the greatest stiffness. As \text{$\alpha$} increases (e.g., $\text{$\alpha$} = 4, 8$), the RF-Strain curve exhibits a gentler slope. At larger taper angles, the RF-Strain curves exhibit a significant downward trend. Changing the taper angle alters both the geometric design and force distribution of the panel, thus affecting its mechanical properties. The tapered structure may redistribute the external load by reducing its vertical component and increasing the horizontal component, resulting in more lateral deformation.


\subsubsection{Crease thickness \( t_c \)}
We kept the facet thickness constant at 1.5 mm while varying the crease thickness \( t_c \) (M12 to M16), and measured the corresponding RF-strain curves. The results indicate that as crease thickness increases relative to facet thickness, the bending stiffness of the crease rises significantly (Fig. \ref{fig:simulation results}c). By tuning the relative thickness between facets and creases, we can program the panels to have different compliance and mechanical properties.

\subsubsection{Scaling ratio n}
Finally, we used M17 as the baseline model and investigate the effect of size scaling ratios (M17 to M22). The output force in simulations maintains a linear relationship with the scaling factor (Fig. \ref{fig:simulation results}d). This indicates that specific scaling factors can be selected based on performance requirements, enabling adjustable mechanical properties to suit different applications.

Overall, our design and analysis deliver a tunable and constant force output in a targeted folding mode of an origami-inspired mechanism. We took the reaction force at a strain of 0.25 as the baseline, and can obtain the range of constant force where the magnitude fluctuates within 5\% (Fig. \ref{fig:simulation results}d). The constant force output covering most of the compression process provides a theoretical principle for applying origami-inspired structures in soft grippers.


\section{Experiment and Analysis}

\subsection{Module Fabrication}

\begin{figure}[!t]
    \centering
    \includegraphics[width=0.9\linewidth]{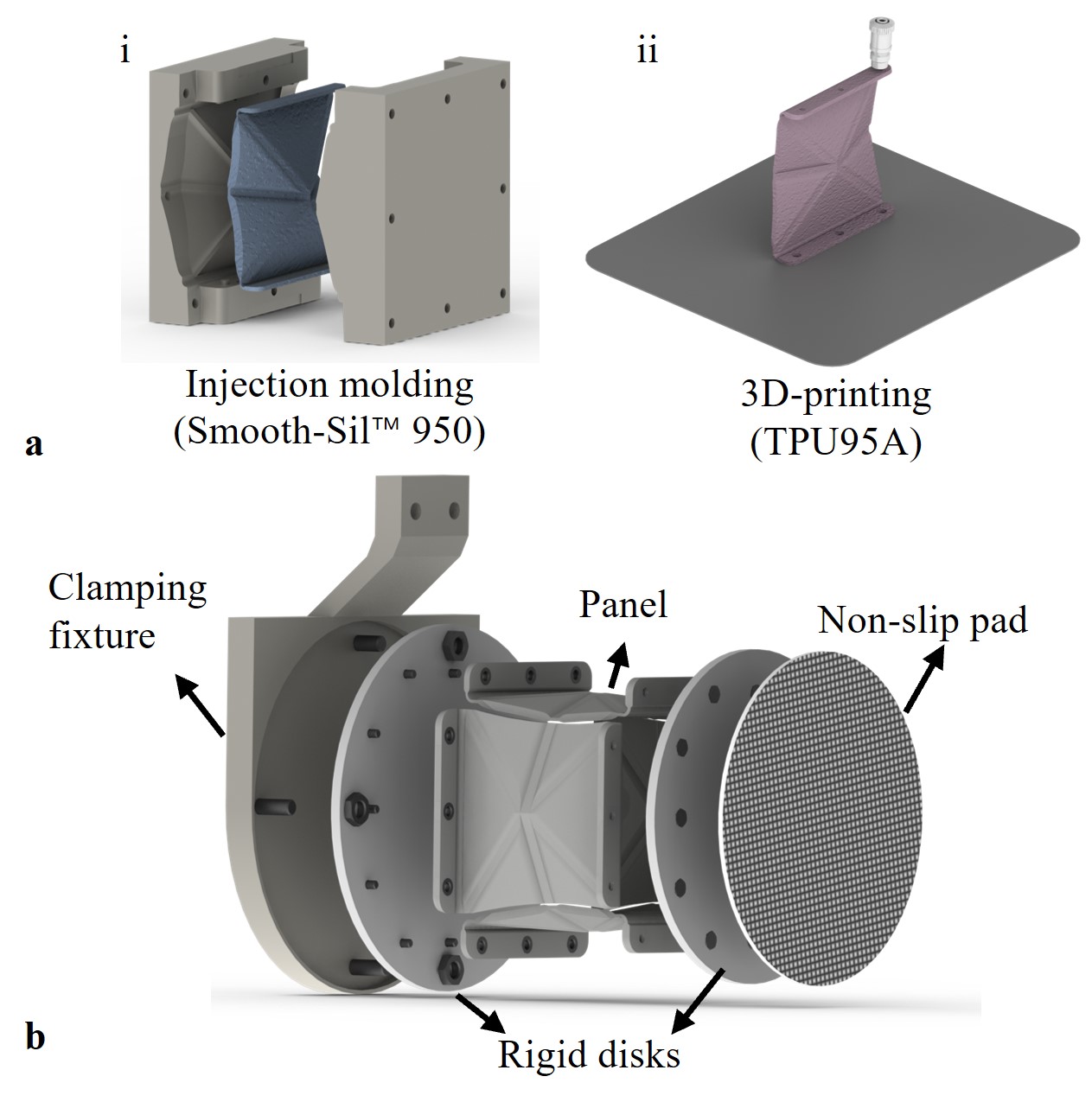}
    \caption{Fabrication and assembly of the origami-inspired gripper.~(a) Panel fabrication by (i)~silicone casting and (ii) TPU 3D printing.
    ~(b)~Assembly of the gripper.}
    \label{fig:fabrication and assembly}
\end{figure}

The experimental module was fabricated by using the simulated model with scaling ratio \textit{n} = 0.6. We fabricated TUP-based (3D-printed in TPU 95A HF by Bambu Lab X1) and silicone-based (Mold casting with Smooth-Sil™ 950) panels. Each silicone panel was manufactured via injection molding (Fig. \ref{fig:fabrication and assembly}a). The soft panels and rigid disks (3D-printed in PLA) are assembled with the similar setup shown in simulations (Fig. \ref{fig:fea}c). Each panel was positioned at a radial distance of 20 mm. The end and base disks have diameters of 62 mm and 65 mm, respectively. The assembled module can be used as a soft gripper with a clamping fixture (Fig. \ref{fig:fabrication and assembly}b) connected to a robotic arm (Franka Emika robot).

Three distinct panel models, Ma, Mb, and Mc, were tested with varying geometric parameters. Ma has a \(dist\) of 2.25~mm and a taper angle of 3°, whereas Mb features a \(dist\) of 3.00~mm and a taper angle of 4°. Both Ma and Mb were fabricated using TPU, while Mc was manufactured using Smooth-Sil 950 with a \(dist\) of 4.50~mm and a taper angle of 6°. Other common parameters include a lower base length (\( l_1 \)) of 37.5 mm, an upper base length (\( l_2 \)) of 33.75 mm, a module height (\( H_0 \)) of 32.5 mm, a facet thickness (\( t_f \)) of 1.5 mm, and a crease thickness (\( t_c \)) of 0.54 mm. 
 


\subsection{Compression Experiments}

\begin{figure}[!t]
    \centering
    \includegraphics[width=0.9\columnwidth]{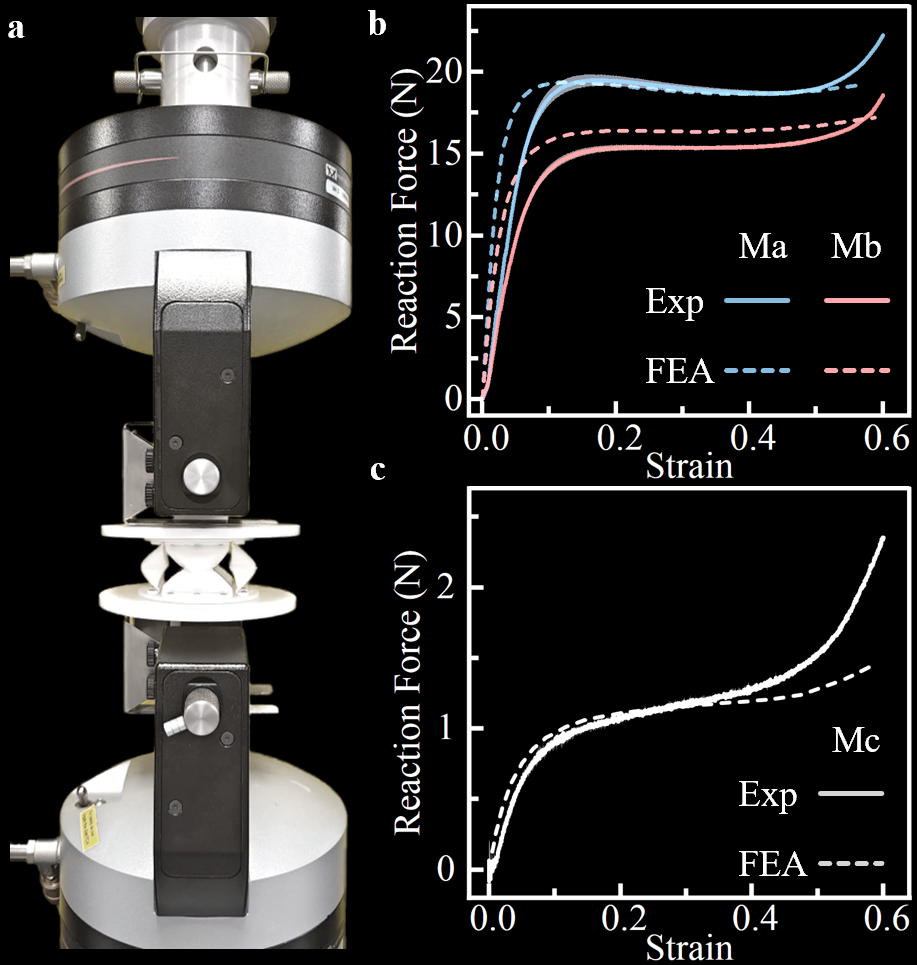}
    \caption{Compression experiments and FEA in (a) TPU-based module (Ma, Mb) and (c) silicone-based module (Mc).
}
    \label{fig:compression test}
\end{figure}

The compression experiments were performed using a universal testing machine fixed with the base and end disk of the module (Fig.~\ref{fig:compression test}a). The loading device compressed the structure at a rate of 2.5~mm/min. Compression was halted when the origami-inspired module reached 60\% of its initial height (20~mm), at which point the entire structure was nearly fully folded. Displacement and load data were recorded using a data acquisition system.

Experimental results and FEA simulations of three modules (Ma, Mb, Mc) were compared as shown in Fig.~\ref{fig:compression test}. The constant force outputs of Ma and Mb were approximately 15~N and 20~N, respectively, with both maintaining a stable force over a strain range from 0.1 to 0.6. Ma exhibits a higher constant force output compared to Mb. In contrast, Mc produced a constant force output of approximately 1.1~N, significantly lower than that of Ma and Mb. Beyond the constant force range, all three modules exhibited an increasing force trend, indicating that the structures were fully folded and the compressed surfaces were in contact, thereby generating additional pressure. Overall, the simulation results are close to the experimental tests, which proves the feasibility and reliability of our simulation method and designed structure.

\subsection{Torsional Stiffness Experiments}

\begin{figure}[!t]
    \centering
    \includegraphics[width=1\columnwidth]{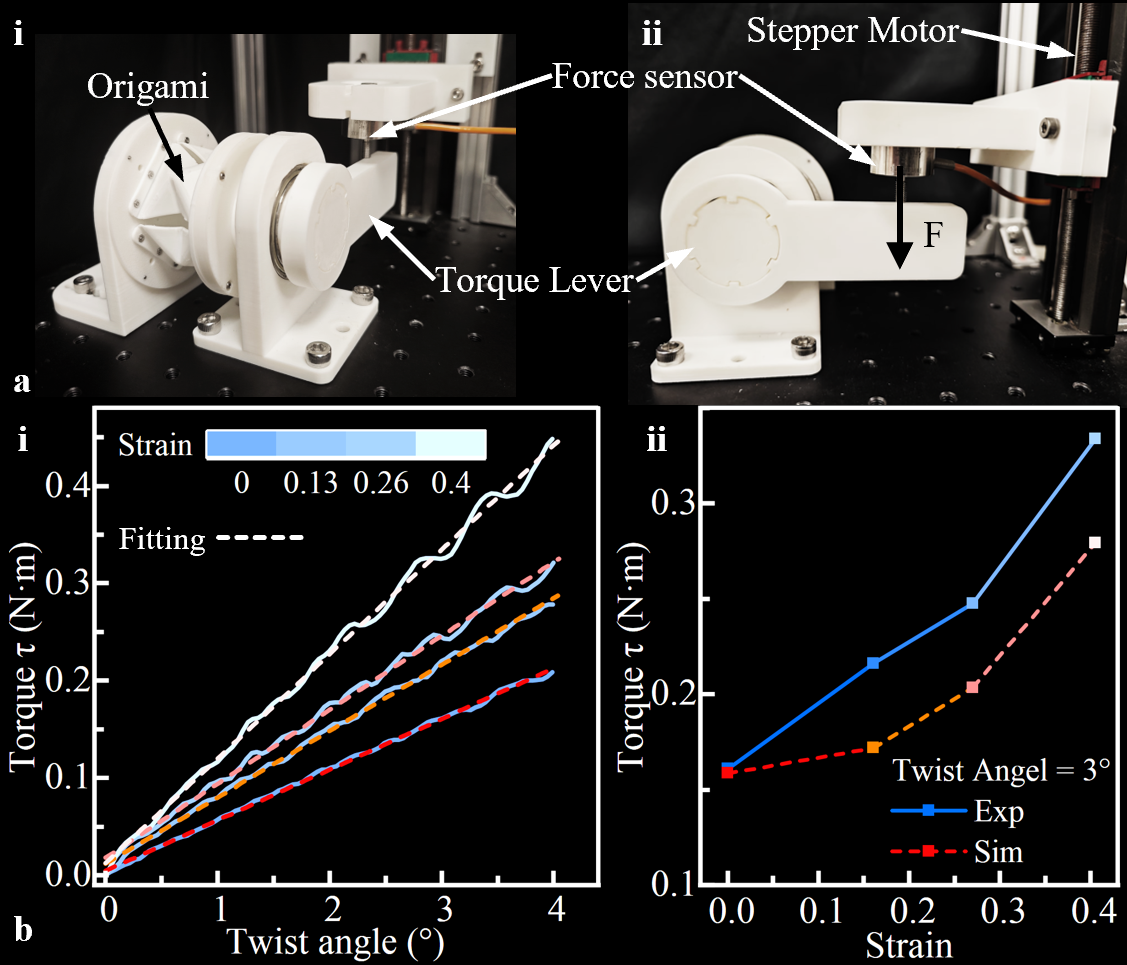}
    \caption{Torsional experiment. (a) Test platform. (b) Experimental results showing (i) torque versus twist angle at different strains with fitted curves and (ii) experimental and simulated torque-strain relationships at a twist angle of $3^\circ$.}
    \label{fig:Twisting test}
\end{figure}

To evaluate the torsional stiffness of the origami-inspired structure, we build a measuring platform (Fig. \ref{fig:Twisting test}a). The base disk was fully constrained but able to be fixed at different positions, while the top was fixed to the rotating fixture, allowing for controlled twisting deformation. A lever arm was rigidly attached to the rotating structure and extended radially from its central axis. The other side of the lever arm was pushed by a linear stepper motor, inducing a rotational torque $\tau$ in the module.

A force sensor was placed between the motor and the lever arm to measure the pushing force F. The applied torque was determined as \(\tau = \mathrm{F} \times \mathrm{L}\), where L is the distance from the rotational center to the applied force. The bottom of the module was fixed at various positions to achieve different strains, and the motor moved downward slowly to maximum displacement (maximum twisting angle around 4°). 

The experimental and simulation results indicate that the origami-inspired module exhibits high twisting stiffness, with torque increasing as strain rises (Fig. \ref{fig:Twisting test}b). Simulations with the same configuration were also performed to compare the experimental stiffness.
High torsional stiffness ($>$ 0.1 N$\cdot$m/°) enhances the gripper's resistance to external disturbances, reduces torsional deformation, and ensures stable operation even in dynamic environments.

\subsection{Grasping Experiments}

\begin{figure}[!t]
    \centering
    \includegraphics[width=1\columnwidth]{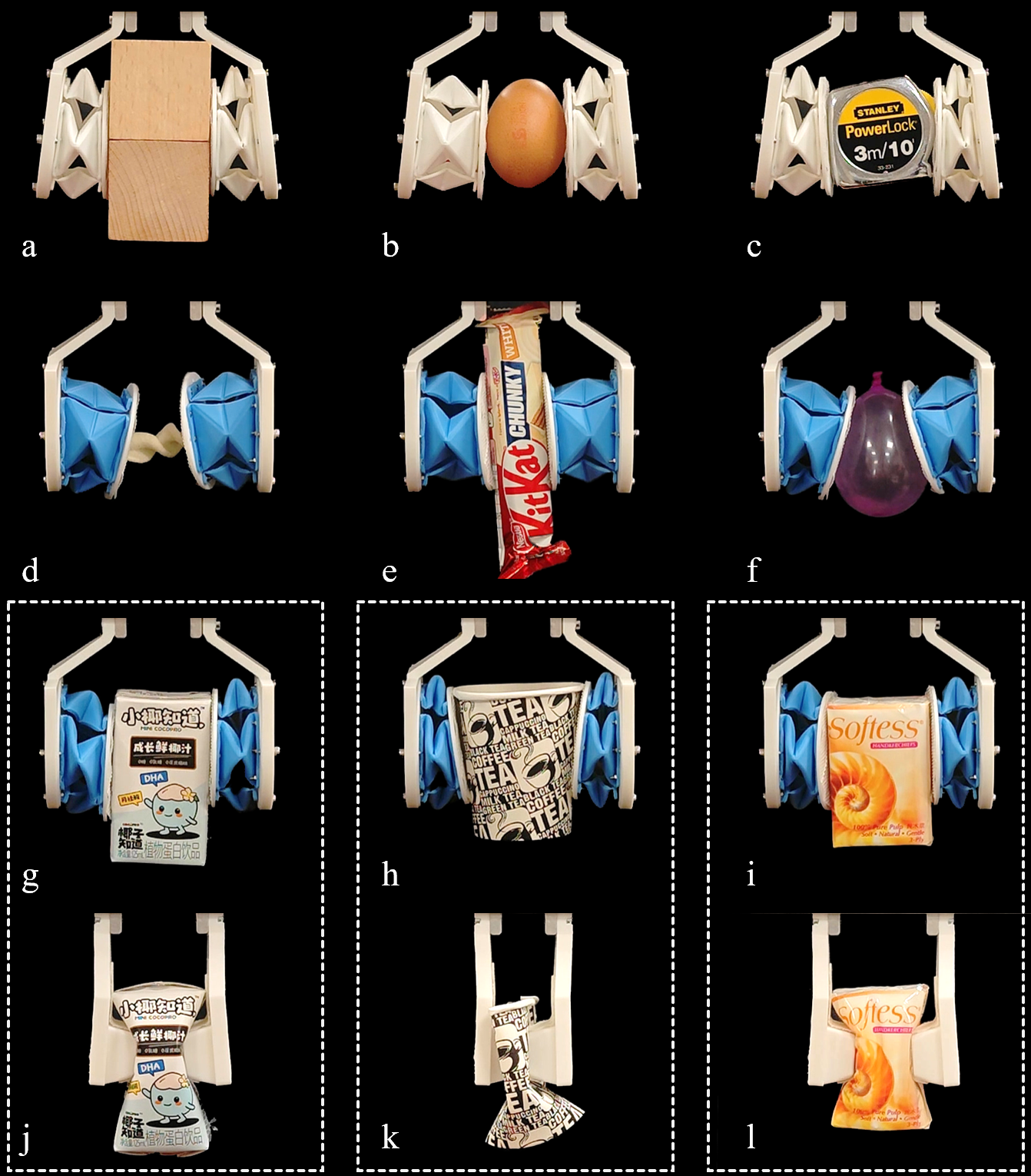}
    \caption{Grasping experiments.  
(a)–(c) TPU-based gripper (a wood block, an egg, and a tape measure).  
(d)–(f) Silicone-based gripper (a snack, a chocolate, and a balloon).  
(g)–(l) Comparative demonstrations with a purely rigid gripper.}
    \label{fig:Grasping experiments}
\end{figure}

We demonstrated and validated the grasping performance of the origami-inspired structure (Model Ma and Mc). 
Owing to differences in materials and the magnitude of constant force, the TPU-based gripper was employed for heavier and harder objects, whereas the silicone-based gripper was used for lighter ones. 
As shown in Fig.~\ref{fig:Grasping experiments}, both grippers successfully grasped various objects, demonstrating stable holding performance that reflects their constant force output capability. These results indicate that the soft material and structures of the origami-inspired grippers can be optimized and assembled to suit target objects.

We also compared the grasp function and adaptability between origami-inspired and purely rigid grippers. 
The results show that our soft gripper can gently deform into fragile objects due to the inherent structural properties (softness and constant force output), while the rigid gripper crushed the object without precise force control. Our method prevents excessive force application on fragile objects, reducing the risk of damage, which is particularly useful for grippers without an integrated feedback system.

Additionally, our gripper demonstrated the ability to dynamically conform to the shape of an object. For example, when grasping a paper cup, the gripper naturally conforms to the outer surface of the cup, increasing the contact area and thus enhancing friction and achieving a more stable grasp.

\section{CONCLUSIONS}
In this work, we presented an origami-inspired soft gripper design capable of a constant force output over a wide deformation range. 
Our FEA simulations and experimental results consistently demonstrated the effect of geometry parameters, including protrusion distance \textit{dist}, taper angle \( \alpha \), and crease thickness \( t_c \) on the intrinsic mechanical properties. 
By exploiting the waterbomb-like folding mode, the gripper achieves stable and predictable force distribution, simplifying precise force control and reducing damage to delicate objects.

Beyond verifying the feasibility of constant-force grasping through both compression and twisting tests, our grasping demonstrations highlighted the structure’s inherent compliance, adaptability to various object shapes, and improved safety compared to purely rigid grippers. These promising results point to significant potential for the origami-inspired approach in soft robotics applications such as industrial automation, food handling, and human–robot interaction.

\section*{ACKNOWLEDGMENT}
We thank Xiyang Zhu, Guanglu Jia, Rui Peng, and Xiaolei Wang for fruitful discussions and suggestions.

		\bibliographystyle{IEEEtran}
		\bibliography{citationlist}

\end{document}